\documentclass[cameraready]{Interspeech}

\title{Joint Speech And Text Training For LLM-based End-To-End Spoken Dialogue State Tracking}

\author[affiliation={1}, orcid=0009-0009-1659-0459, equalcontribution]{Katia}{Vendrame}
\author[affiliation={1}, orcid=0000-0001-9852-3456, equalcontribution]{Bolaji}{Yusuf}
\author[affiliation={1}, orcid=0000-0002-3725-742X]{Santosh}{Kesiraju}
\author[affiliation={1}, orcid=0009-0005-9875-3012]{Šimon}{Sedláček}
\author[affiliation={1}, orcid=0000-0001-7938-3945
]{Oldřich}{Plchot}
\author[affiliation={1}, orcid=0000-0002-8800-0210]{Honza}{Černocký}

\address{
    $^1$ Speech@FIT, Brno University of Technology, Czechia 
}

\email{ivendrame@fit.vut.cz, iyusuf@fit.vut.cz}


\usepackage{comment}
\usepackage{amsmath,graphicx,hyperref,booktabs,multirow,xcolor,pifont}
\usepackage{mystyle}
\usepackage{subcaption}
\usepackage{float}
\usepackage[table]{xcolor}

\begin{document}

\maketitle

\begin{abstract}

End-to-end spoken dialogue state tracking (DST) is made difficult by having to handle speech input with scarce training data.
Combining speech foundation encoders and large language models has emerged as an approach to alleviate this difficulty.
Although this approach results in strong spoken DST models, achieving state-of-the-art performance in realistic multi-turn DST, it struggles to generalize across domains and requires annotated spoken DST training data for each domain of interest.
However, collecting such data for every target domain is both costly and difficult.
Noting that textual DST data is more easily obtained for various domains, in this work, we propose jointly training on available spoken DST data and written textual data from other domains as a way to achieve cross-domain generalization.
We conduct experiments which show the efficacy of our proposed method for getting good cross-domain DST performance without relying on spoken training data from the target domains.
\end{abstract}

\section{Introduction}
\label{sec:intro}



Task-oriented dialogue (ToD) systems are conversational agents that assist users in achieving certain defined goals such as making reservations, locating attractions, requesting information, etc.
As the dialogues typically comprise multiple user-agent turns in unstructured natural language, a key component of ToD systems is dialogue state tracking (DST)---the extraction of structured domains and slots from the natural language conversation, e.g., from conversation turn: \textit{``It should leave after 13:45."} and its context, it should extract the following domain and slots: \textit{\{train: \{day: Sunday, departure: Cambridge, leave at: 13:45\}\}}.

That speech constitutes a natural and often more convenient medium of communication has prompted research in DST from spoken rather than text input~\cite{soltau-etal-2023-dstc}.
This conventionally entailed cascading an automatic speech recognition (ASR) module, optional error correction modules, and text-based DST modules~\cite{jiang-etal-2023-speech,yoon-etal-2023-adapting}.
However, recent end-to-end (E2E) methods have been developed which are trained to directly predict the dialogue states from speech~\cite{ganhotra21_interspeech,sunder23_interspeech,wang2024retrieval,sedlacek25_interspeech}.

E2E methods alleviate some of the deficiencies of their cascade counterparts such as system complexity, error propagation, latency, loss of paralinguistic information and other discrepancies between spoken and written language.
However, they suffer from a data scarcity problem---collecting enough spoken dialogue data for training a well-performing spoken DST model from scratch is a laborious and expensive undertaking.

A recent E2E method~\cite{sedlacek25_interspeech}---which we take as a baseline in this work---tackles the data scarcity problem by using a connector module to link a pretrained speech encoder and a pretrained large language model (LLM) combining the former's robust speech modeling ability with the latter's ability to extract complex structured information from natural language.
The connector is then trained along with low-rank adapter (LoRA)~\cite{hu2022lora} layers to directly predict the dialogue states encoded as a JSON string from the speech input into the encoder.
While this approach proved successful, considerably outperforming previous state-of-the-art DST systems on the multi-domain SpokenWOZ~\cite{si2023spokenwoz}, it struggles to generalize to domains (specifically, slot values corresponding to names of places, restaurants, etc.) other than those on which it was trained.

Given the cost and difficulty of collecting spoken DST training data for every target domain, it is necessary to develop methods that generalize to multiple domains without access to spoken training data from those domains.
We note that \textit{unpaired}\footnote{Note that by unpaired, here and throughout the paper, we mean DST data with no spoken utterances as opposed to text data with no DST labels.} textual DST training data is more abundant and less costly to collect.
Moreover, prior work has shown that slot-augmentation can significantly improve the performance and robustness of text-based DST systems~\cite{tian2021tod,thulke2021adapting}.
Therefore, inspired by similar joint training methods that have been used for other speech processing tasks 
\cite{tang-etal-2021-improving,chen22r_interspeech,yusuf2022usted,thomas2022towards,yusuf2024written,dao25_interspeech}, we propose a joint training method combining single-domain spoken DST data with cross- or multi-domain textual data for training\footnote{We released code and data at \url{https://github.com/kackav/dialogue_state_tracking}}.


In our method, we augment the LLM-based E2E speech-to-DST model with a text encoder whose input is the natural language query from the user and whose output is passed into the connector and then into the LLM to predict the dialogue states.
By sharing parameters (specifically the connector and the LoRA layers in the LLM) between the speech and text pipelines and jointly training on spoken DST data from some source domain(s) and textual DST data from target domains, we aim to learn a multi-domain model capable of conducting DST in the target domain.

We conduct experiments on the SpokenWOZ~\cite{si2023spokenwoz} and Speech-Aware MultiWOZ~\cite{soltau23_interspeech} datasets, where we train on speech data from a \textit{source} domain jointly with text from a \textit{target} domain.
These experiments show that the proposed method consistently improves cross-domain generalization across datasets and models, including scenarios where the target text forms only a small portion of some larger text corpora used for training. Furthermore, we compare this approach to an alternative strategy that converts annotated user turns into speech using a Text-To-Speech system (TTS), and demonstrate that our proposed method achieves comparable results, without the added complexity of TTS---an important advantage for low-resource languages where high-quality TTS systems are often unavailable.
\begin{figure}[t!]
\centerline{\includegraphics[width=0.5\columnwidth,angle=270]{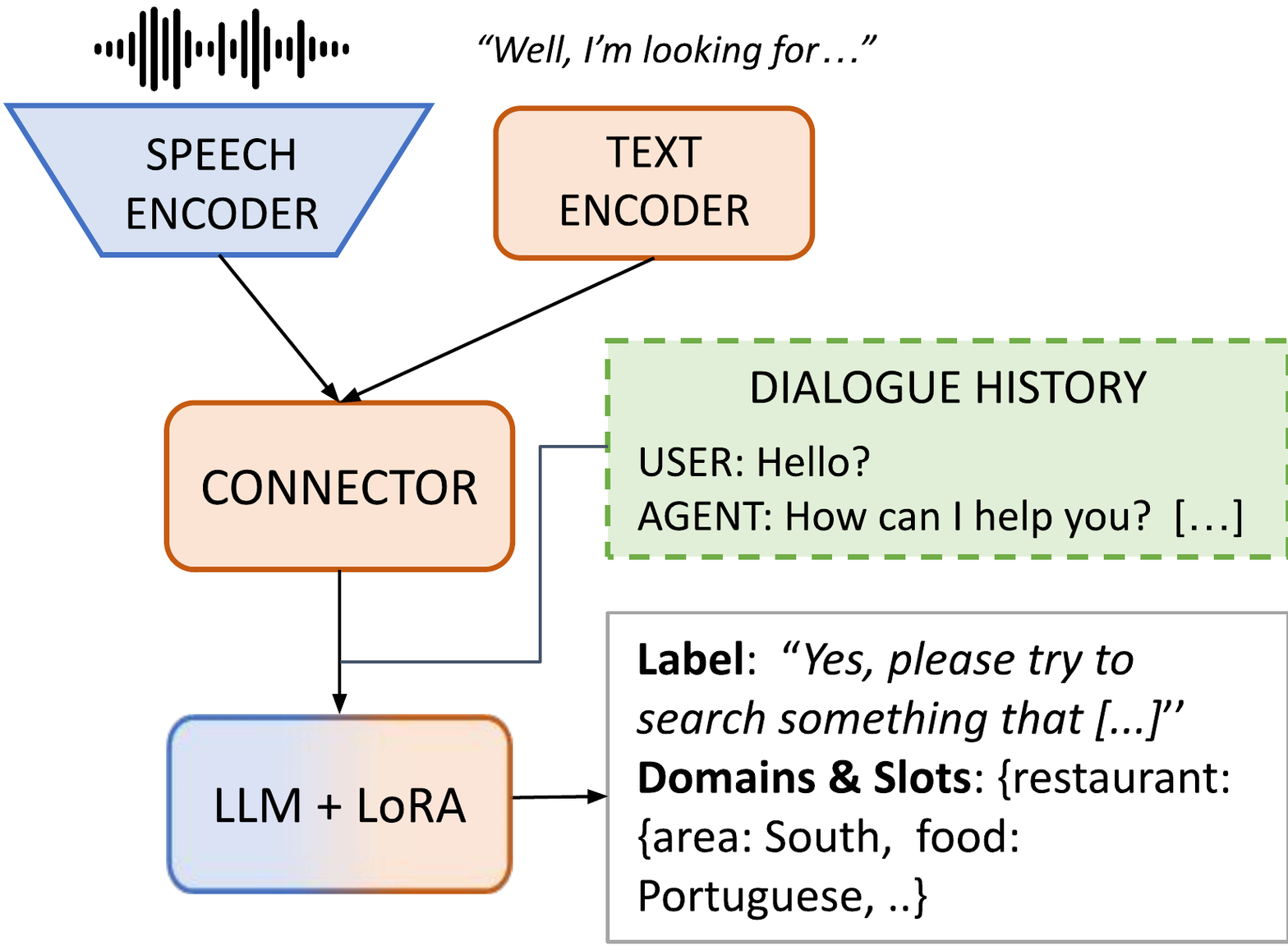}}
  \caption{Joint speech and text E2E DST model.}
  \label{scheme}
\end{figure}
\subsection{Methods}
\label{sec:methods}

Our DST system is based on the E2E framework from~\cite{sedlacek25_interspeech}, which uses a model comprising a speech encoder, a large language model and a connector module used to transform the outputs of the former into the embedding space of the latter (illustrated in Figure~\ref{scheme}).
The model's inputs are the speech corresponding to the user input at the current turn of the dialogue and the dialogue history in the form of transcriptions of user and agent turns of the dialogue with a new-line character (\texttt{\textbackslash n}) inserted between turns.
The speech is encoded by the tandem of the encoder and the connector, and resulting speech embeddings are prepended to the LM embeddings of the dialogue history.
The resulting sequence is passed through the LLM transformer layers.
The model is trained to output a single JSON string encoding a Python dictionary that contains the ASR transcription and the dialogue states, i.e., the set of active domains and corresponding slots.



\subsection{Joint Speech and Text E2E DST (Multimodal DST)}

We propose modifying the E2E DST model to accommodate training on not just spoken but also written task-oriented dialogues.
To this end, we augment it with a text encoder which accepts as input the natural language user input of the current user turn of the dialogue.
As with the speech encoder output, the text encoder output then goes through the connector, is prepended to the dialogue history and then fed into the language model to predict dialogue states.

The inclusion of the text encoder allows us to train on diverse text DST data without requiring paired spoken user utterances.
We hypothesize that this would allow us to improve the model performance on spoken data from those domains, without needing to synthetically generate them.
Note that the text encoder is only used to aid training and can be discarded during inference, thus incurring no additional inference cost compared to the baseline E2E DST system.




\subsection{Training}

Following the findings of~\cite{sedlacek25_interspeech}, we train the model in two phases:
First, we freeze the LM and pretrain the encoder and connector for ASR on diverse training data using the cross-entropy objective.
Then, we introduce the text encoder and insert LoRA layers into the LM, and finetune the connector, text encoder and LoRA parameters jointly for DST from speech and text with the speech encoder and base language model frozen.

At each step of training, we feed a speech batch and a text batch through their respective encoders and train to minimize a sum of three cross-entropy losses.
In addition to the cross-entropy objectives computed from the DST on speech data and text DST on unpaired text data, we also train the model for text DST from the transcription of the speech batch.

\section{Experiments}
\label{sec:experiments}


\subsection{Datasets and metrics}
\label{section:experimente:datasets}
We conduct our experiments on two spoken dialogue state tracking datasets: SpokenWOZ (SW)~\cite{si2023spokenwoz} and Speech-aware MultiWOZ (MW)~\cite{soltau23_interspeech}.
In most of our experiments, we utilize a training set from one of them as the paired speech-text data and the other as unpaired text data, e.g., training on a combination of SpokenWOZ speech and MultiWOZ text and vice versa, allowing us to measure the cross-domain performance.
We also use text from DialogStudio (DS)~\cite{zhang-etal-2024-dialogstudio} as an alternative unpaired text to measure the impact of training on unpaired DST data, which does not exactly match the target domain.
For SpokenWOZ, we follow the pre-processing from~\cite{sedlacek25_interspeech}, discarding nine corrupted conversations from the SpokenWOZ test sets, and employing Whisper-large-v3~\cite{radford2023robust} to re-transcribe the user and agent speech.
We note that MultiWOZ presents an additional challenge because its validation and test sets contain data from a different city (New York) than the training set (Cambridge); therefore, we expect using MultiWOZ training text for joint training would improve the slot keys but not necessarily the values for the MultiWOZ.

For ASR training, we use the Fisher~\cite{cieri2004fisher}, Librispeech~\cite{panayatov2015librispeech}, CommonVoice (version 17)~\cite{ardila-etal-2020-common} and VoxPopuli~\cite{wang2021voxpopuli} datasets.

We report the joint goal accuracy (JGA) computed using the MultiWOZ evaluation script\footnote{github.com/Tomiinek/MultiWOZ\_Evaluation}~\cite{nekvinda-dusek-2021-shades}.
As is standard practice in DST, before scoring, we post-process the outputs of all DST systems using fuzzy matching\footnote{pypi.org/project/fuzzywuzzy/} to match hypothesized slots to the closest slots by edit distance in the ontology.


\subsection{Model architecture and hyper-parameters}


We use WavLM~\cite{chen2022wavlm} as the pretrained speech encoder in all our experiments, and Gemma-3-1B-it~\cite{gemma_2025} as the LLM in most experiments.
We also report a subset of results with Gemma-3-4B-it and Gemma-3-12B-it to quantify the impact of scaling, as well as OLMo-1B~\cite{Groeneveld2023OLMo_} to facilitate comparison with prior state-of-the-art~\cite{sedlacek25_interspeech} and because the openness of its training data and configuration allay risks of data contamination.

For the connector, we use a 4-layer Transformer encoder with hidden size 1024, 4 attention heads, feedforward intermediate dimensions of 4096 and sinusoidal positional encoding with maximum length set to 512.
The Transformer layers are preceded by two convolutional layers with strides of 3 and 2, so that the speech input to the LLM has 8 Hz.
Note that text inputs from the text encoder are passed through only the Transformer and not the convolutional module of the connector.

For the text encoder, we use a Transformer encoder with identical dimensions to the connector Transformer, preceded by an input embedding layer with the same tokenizer as the respective LM.
To compare the text encoder pipeline with training on synthetic speech, we generated spoken user turns using Qwen3-TTS-12Hz-0.6B-CustomVoice\footnote{The voice was generated choosing the speaker ``Ryan" from the model's speaker list}~\cite{qwen}.


In the ASR training phase, we employ AdamW optimizer~\cite{loshchilov2018decoupled} for 100k steps. 
The learning rate is warmed up over the first 1000 steps up to a peak of $2\times10^{-4}$ and then decayed linearly to $2\times10^{-6}$ over the course of training.

In the finetuning phase, we use LoRA with rank and alpha 32 and train using AdamW.
We warm up the learning rate to a peak of $5\times10^{-5}$ over 1000 steps, and then decay it for up to 60k steps.
We measure the JGA (in teacher-forcing mode) on the validation sets and use it for early stopping and for selecting the best checkpoint for each validation set.






\section{Results}
\subsection{Training with text data from various sources}

\begin{table}[t!]
\centering
\caption{Joint goal accuracy (\%) as unpaired text data is varied on E2E model with Gemma-1B-it.}\label{gemma1_results}
\resizebox{0.79657\columnwidth}{!}{%
\begin{tabular}{ccccc}
\toprule
&{Training } & {Training} & {SW} & {MW} \\
&Speech & Text & Val & Val\\
\midrule
A1&\multirow{5}{*}{SW}  & \ding{55}     & 36.1 & \xdomain{15.1} \\
\cmidrule(lr){3-5}
A2 && MW                         & 36.3 & \xdomain{19.0} \\
A3& & MW, DS    & 37.9 & \xdomain{18.4} \\
A4 & & DS  &      37.1 & \xdomain{17.1}\\           
\midrule
B1 & \multirow{5}{*}{ MW}   & \ding{55}     & \xdomain{20.5}& 28.6 \\
\cmidrule(lr){3-5}
B2 & & SW                            & \xdomain{30.6} & 24.7\\
B3 & & SW, DS                        & \xdomain{28.8} & 23.6 \\
B4& &DS                           &\xdomain{20.5} &28.1 \\   
\midrule
C1 & SW & MW-train+test & 38.7 & 23.1 \\
C2 & MW & SW-train+test & 32.6 & 25.7 \\
C3 &MW+SW   & \ding{55}  & 36.2 & 16.9 \\
C4 &MW+SW   & MW-val, SW-val   & 55.3 & 42.4 \\
\bottomrule
\end{tabular}%
}
\end{table}

First, we explore the impact of training with text alongside speech while varying the degree of relevance to the target test sets, and we report these in Table~\ref{gemma1_results}.

Comparing rows A1 to A2, we find that training on paired SpokenWOZ data with unpaired MultiWOZ training text considerably improves performance on the MultiWOZ validation set compared to the model trained with no text, reducing the JGA gap to the model trained on paired MultiWOZ data by 28.9\%.
In the converse case, comparing B1 to B2, we observe even larger improvements with a 64.7\% reduction of the gap.

Next, we consider the case where the target text is mixed with DialogStudio text (rows A3 and B3).
This emulates a setting where a large text corpus, including text from the target domain, is available for training, but the target domain is not necessarily known a priori.
This slightly degrades the target set performance compared to using only target training data as the paired text, but is still considerably better than the baseline trained with no unpaired text.

Subsequently, we train using only unpaired text from DialogStudio without text from the target domain (rows A4 and B4), and observe moderate improvement over the baseline on the MultiWOZ validation set but none on SpokenWOZ.
This can be explained by the fact that the DialogStudio collection contains the MultiWOZ training data but not SpokenWOZ.

Overall, these results indicate that our proposed method enables adaptation to a target domain using only unpaired text from that domain, even when it forms only a small portion of a larger text corpus.

Finally, we report results on a set of toplines.
In C1 and C2, we use as unpaired text the concatenation of the train and test sets of the target domain (while still testing on the validation set).
We note that while C2 slightly outperforms B2 on SpokenWOZ (6.5\%), C1 yields a more substantial relative improvement of 21.6\%.
This latter result shows the impact of training on text data with not just overlapping the domains but also the slot values since the MultiWOZ test and validation sets contain dialogues from the same city.
In C3, we train on paired data from both SpokenWOZ and MultiWOZ, and in C4, we additionally use the text from the \textit{validation} sets as unpaired texts.
C4 massively outperforms C3 even if the former is already trained on speech data from both domains.
Although this last experiment is only an oracle one, it underscores the capability of our method to adapt under idealized conditions.

\subsection{Mechanism of incorporating unpaired text}
While seeing the benefits of training on unpaired text, a natural question arises as to whether and how much the text encoder actually contributes.
Specifically, if the improvements are a result of simply exposing the LM to dialogue states from the target domain, we may simply finetune the LoRA parameters on the unpaired data without a text encoder at all. To appropriately contextualize these two results, we compare these results with TTS-generated speech (Table~\ref{tab:TTSandTE}).
\par Training without a text encoder (D1 and D3) performs worse on the target domain than with it (A2 and B2, respectively), but consistently improves upon the baseline with no unpaired text (A1 and B1).
This indicates that while part of our improvements stem from simply exposing the LM to the target domain dialogue states, another part of it comes from improving the speech pipeline's ability to extract these dialogue states from natural language input.
\par Using synthetic speech further recovers the JGA gap between training on the target domain speech and training on target text data by 49\% in the case of SpokenWOZ (D4) and by 27\% for MultiWOZ speech (D2), by slightly losing performance on the original speech domain. 
Nonetheless, by appropriately tuning the weight of the text loss, we can trade in-domain performance for target-domain gains and obtain similar results to those achieved with TTS. 
Specifically, setting the weight of the text loss to 8, on the target domain task we achieve 32.4\% JGA on SpokenWOZ and 21.8\% on MultiWOZ. Joint text training thereby constitutes a valid alternative when high quality TTS systems are unavailable. 



\begin{table}[t]
\centering
\caption{Joint goal accuracy (\%) on Gemma-1B-it with or without text encoder or with TTS-generated data.}
\label{tab:TTSandTE}
\resizebox{0.93\columnwidth}{!}{%
\begin{tabular}{ccc cccc}
\toprule
&{Training} & {Training}&TTS    &Text       & {SW}  & {MW} \\
& Speech    & Text      &Speech & Encoder   &Val    &Val\\
\midrule
A1&\multirow{5}{*}{SW}&\ding{55}&\ding{55} & \ding{55}     & 36.1 & 15.1 \\
\cmidrule(lr){3-7}
{A2}&&  \multirow{2}{*}{MW}   &\ding{55}   &\ding{51}& 36.3 & 19.0 \\
D1&         &                 &\ding{55} &\ding{55} &36.1&17.4\\
\cmidrule(lr){3-7}
D2&    &\ding{55}  & MW  &\ding{55}& 33.3 & 21.6\\
\midrule
B1 & \multirow{5}{*}{MW}  &\ding{55}  &\ding{55} & \ding{55}     & 20.5& 28.6 \\
\cmidrule(lr){3-7}
{B2}&&  \multirow{2}{*}{SW}     &\ding{55}&\ding{51}& 30.6 & 24.7\\
D3&         &                  &\ding{55} &\ding{55} & 28.6&24.9\\
\cmidrule(lr){3-7}
D4&          &\ding{55}         & SW  &\ding{55} & 33.3&23.6\\
\bottomrule
\end{tabular}
}
\end{table}



\subsection{Test set performance}
In addition to the \mgone{} model used in all experiments above, we also report results obtained with \moone{}, \mgfour{} and \mgtwelve{} on the test sets in Table~\ref{gemmas}.
\moone{} lets us observe how well our method works on an LM from a different family and to compare with~\cite{sedlacek25_interspeech} where most results were from \moone{}.
Experiments on \mgfour{} and \mgtwelve{} show how well our approach translates to larger and better language models. 
In these experiments, we keep the same LoRA rank of 32.

On \moone{}, our implementation of the no-text baselines outperforms the results reported in~\cite{sedlacek25_interspeech} (compare E0 and F0 to E5 and F5, respectively).
We credit this to using more data for the ASR pretraining phase.
Nevertheless, as in the \mgone{} experiments, joint training with target domain text leads to a much improved target domain performance with SpokenWOZ, especially benefiting from target text training. 
Specifically, joint training with SpokenWOZ speech and MultiWOZ text (E6) closes 46\% of the MultiWOZ performance gap between the baseline trained on MultiWOZ speech with no text (F5) and that trained on SpokenWOZ speech (E5).
In the converse experiment, where we train with MultiWOZ speech and SpokenWOZ text (F6), we observe that we can recover 79\% of the performance gap on SpokenWOZ between training on MultiWOZ only (F5) and actual SpokenWOZ speech (E5).

Unsurprisingly, we find that larger LMs generally perform better.
More interestingly, the relative impact of joint text training is more pronounced the larger the LM is.
In the case of the largest model in our experiment, \mgtwelve{}, joint training with SpokenWOZ text and MultiWOZ speech (F4) yields almost identical SpokenWOZ test set JGA as training on SpokenWOZ speech (E3).
Similarly, we observe larger relative improvements on MultiWOZ with the larger models.
Moreover, in the case of MultiWOZ, where adding SpokenWOZ text degrades the MultiWOZ test set JGA for \mgone{} (B1 vs B2), we observe that this performance degradation set decreases (F3 vs F4) or disappears (F1 vs F2) for the larger models.

As audio-capable large language models are increasingly being adopted for broader speech understanding tasks, we present for comparison results obtained by prompting Gemini-2.5-flash~\cite{gemini25} to transcribe and predict dialog states.
The JGA values obtained (G1) are worse than the in-domain performance of our \mgtwelve{} no-text baseline (E3 and F3), but better than the results on out-of-domain validation sets. Nevertheless, joint-text training produces significantly better in- and out-of-domain results on SpokenWOZ and on MultiWOZ (E4 and F4), while having remarkably lower inference latency. However, this comparison should be treated cautiously due to limited transparency in Gemini's training data and prompt sensitivity.
 \begin{table}[t]
\centering
\caption{Test set joint goal accuracy (\%) with various LLMs.}
\label{gemmas}
\resizebox{0.95\columnwidth}{!}{
\begin{tabular}{lcc ccc}
\toprule
&{Training} & {LLM} &Training&{SW} & {MW} \\
&{Speech} & &Text&Test&Test\\
\midrule
A1&\multirow{11}{*}{SW}&\multirow{2}{*}{\mgone{}}& \ding{55}     & 35.6 & 14.3 \\
{A2}&&     &MW& 36.3 & 18.4 \\
\cmidrule(lr){3-6}
E1  &&\multirow{2}{*}{\mgfour{}{}}   &        \ding{55}  & 41.0 & 16.9 \\
E2  &           &                 &MW    & 40.3 &22.6 \\
\cmidrule(lr){3-6}
E3  &           &\multirow{2}{*}{\mgtwelve} &     \ding{55}     & 42.6&16.6\\
E4 &           &                 &MW     & 43.0 &  23.5\\
\cmidrule(lr){3-6}
E0~\cite{sedlacek25_interspeech} &&&\ding{55}&32.1&-\\
E5 &            &\moone{}         &  \ding{55} & 34.1 & 13.9 \\
E6  &           &                 &MW    & 34.9& 18.2 \\
\midrule
B1 & \multirow{11}{*}{MW}  & \multirow{2}{*}{\mgone{}} & \ding{55}     & 18.9&27.1 \\
{B2}&&  &    SW& 30.9 & 23.5\\
\cmidrule(lr){3-6}
F1  &  &\multirow{2}{*}{\mgfour{}{}}    &          \ding{55}  & 19.8 & 27.3 \\
F2  &           &                 &SW    & 37.4 & 30.8 \\
\cmidrule(lr){3-6}
F3  &           &\multirow{2}{*}{\mgtwelve} &   \ding{55}      & 20.7&32.4\\
F4 &           &                 &SW    &42.2 &31.6 \\
\cmidrule(lr){3-6}
F0~\cite{sedlacek25_interspeech} &&&\ding{55}&-&18.2 \\
F5 &            &\moone{}         &  \ding{55} & 18.7& 23.2\\
F6  &           &                 &SW    & 29.3 &21.9\\
\midrule
G1 &\ding{55}      &Gemini-2.5-flash& \ding{55} & 26.5&20.8\\
\bottomrule
\end{tabular}%
}
\end{table}         

  
\subsection{Slot and values generation accuracy}
Finally, we analyze the slot generation capabilities and how they affect the JGA values. Since the first step of DST is the transcript generation, we evaluate the WERs on the turn transcriptions (Table~\ref{tab:wers}). When adding MultiWOZ text to SpokenWOZ speech training (A2), the WER improves by 35\% relative to the no-text baseline. However, this does not result in a significant improvement on the entity-level WER (E-WER) and on the slot key F1 score, reflecting the difference in the text data, as the MultiWOZ training set contains dialogues from a different city than the validation. 
\par In fact, the higher JGA obtained by adding SpokenWOZ text to MultiWOZ speech training (B1 vs B2) in Table~\ref{gemma1_results} is shown here to be caused by an improvement in slot keys recognition and E-WER, while the overall ASR capabilities do not improve significantly.
\begin{table}[t]
\centering
\caption{Word, Slot key F1, Entity Error Rates and WERs (\%) with Gemma-1B-it.}
\label{tab:wers}
\resizebox{\columnwidth}{!}{%
\begin{tabular}{ccc cc cc cc}
\toprule
&{Training} & {Training}&  \multicolumn{2}{c}{WER}& \multicolumn{2}{c}{E-WER} & \multicolumn{2}{c}{Slot-key-F1}\\
& Speech & Text & {SW} & {MW}& {SW} & {MW}& {SW} & {MW}\\
&   &   &Val&Val &Val&Val&Val&Val \\
\midrule
A1&\multirow{2}{*}{SW}&\ding{55}& 15.5 & 15.7&12.6&47.1  & 81.8 &74.5\\
A2&&  MW   & 14.3 & 10.3 &10.3&46.1  &81.9 & 75.5 \\
\midrule
B1 & \multirow{2}{*}{MW}  &\ding{55}   & 37.4&10.2&18.8&14.8 & 73.3&79.2\\
B2&&  SW&          36.0&10.3&12.6&14.9 &81.4&80.0\\
\bottomrule
\end{tabular}
}
\end{table}

\section{Conclusions}
\label{sec:conlusions}
In this paper, we have proposed a joint speech and text training method for end-to-end spoken dialogue state tracking.
Our method entails augmenting an LLM-based spoken DST model with a text encoder which allows the model to be trained on arbitrary textual DST data.
We show empirically on the SpokenWOZ and MultiWOZ datasets that this method is able to perform spoken DST on domains for which only textual training data is available, even in the presence of other irrelevant training data.
Furthermore, we show the stability of the model across various language model choices, and similar improvement trends even for stronger language models.

The approach we propose enables training multi-domain spoken DST models with speech data from only a subset of domains and only text data from the others.
This work opens the possibility of future work leveraging strategies known to improve textual DST such as paraphrasing and slot augmentation~\cite{tian2021tod,thulke2021adapting} which would otherwise be difficult to implement directly on speech input.
\section{Acknowledgement}
\label{sec:ack} 
This work was supported by the project “On our own: Opportunities and Risks in the Individualization of Society (PRINS) CZ.02.01.01/00/23\_025/0008710”, which is co-financed by the European Union, and by European Union’s Horizon Europe project No. SEP-210943216 "ELOQUENCE". Computing on IT4I supercomputer was supported by MoE through the e-INFRA CZ (ID:90254).
\section{Generative AI Use Disclosure}
Generative AI tools were not used for the writing of this paper.
\bibliographystyle{IEEEtran}
\bibliography{refs}

@inproceedings{sedlacek25_interspeech,
  title     = {{Approaching Dialogue State Tracking via Aligning Speech Encoders and LLMs}},
  author    = {Šimon Sedláček and Bolaji Yusuf and Ján Švec and Pradyoth Hegde and Santosh Kesiraju and Oldřich Plchot and Jan Černocký},
  year      = {2025},
  booktitle = {{Interspeech 2025}},
  pages     = {1748--1752},
  doi       = {10.21437/Interspeech.2025-2764},
  issn      = {2958-1796},
}

@ARTICLE{chen2022wavlm,
  author={Chen, Sanyuan and others},
  journal={IEEE Journal of Selected Topics in Signal Processing}, 
  title={{WavLM: Large-Scale Self-Supervised Pre-Training for Full Stack Speech Processing}}, 
  year={2022},
  volume={16},
  number={6},
  pages={1505-1518},
  doi={10.1109/JSTSP.2022.3188113}}

@article{gemma_2025,
  title={Gemma 3 technical report},
  author={Gemma Team},
  journal={arXiv preprint arXiv:2503.19786},
  year={2025}
}

@article{qwen,
  title={Qwen3-TTS Technical Report},
  author={Hangrui Hu and Xinfa Zhu and Ting He and Dake Guo and Bin Zhang and Xiong Wang and Zhifang Guo and Ziyue Jiang and Hongkun Hao and Zishan Guo and Xinyu Zhang and Pei Zhang and Baosong Yang and Jin Xu and Jingren Zhou and Junyang Lin},
      year={2026},
      eprint={2601.15621},
      archivePrefix={arXiv},
      primaryClass={cs.SD},
      url={https://arxiv.org/abs/2601.15621}, 
}

@article{gemini25,
      title={Gemini 2.5: Pushing the Frontier with Advanced Reasoning, Multimodality, Long Context, and Next Generation Agentic Capabilities}, 
      author={Gheorghe Comanici and Eric Bieber et al.},
      year={2025},
      eprint={2507.06261},
      archivePrefix={arXiv},
      primaryClass={cs.CL},
      url={https://arxiv.org/abs/2507.06261}, 
}

@article{Groeneveld2023OLMo_,
  title={OLMo: Accelerating the Science of Language Models},
  author={Groeneveld, Dirk and others},
  journal={Preprint},
  year={2024}
}

@inproceedings{si2023spokenwoz,
title={Spoken{WOZ}: A Large-Scale Speech-Text Benchmark for Spoken Task-Oriented Dialogue Agents},
author={Shuzheng Si and others},
booktitle={Thirty-seventh Conference on Neural Information Processing Systems Datasets and Benchmarks Track},
year={2023},
url={https://openreview.net/forum?id=viktK3nO5b}
}

@inproceedings{soltau-etal-2023-dstc,
    title = "{DSTC}-11: Speech Aware Task-Oriented Dialog Modeling Track",
    author = "Soltau, Hagen  and
      Shafran, Izhak  and
      Wang, Mingqiu  and
      Rastogi, Abhinav  and
      Han, Wei  and
      Cao, Yuan",
    booktitle = "Proceedings of the Eleventh Dialog System Technology Challenge",
    month = sep,
    year = "2023",
    address = "Prague, Czech Republic",
    url = "https://aclanthology.org/2023.dstc-1.25/",
    pages = "226--234"
}

@inproceedings{soltau23_interspeech,
  title     = {{Speech Aware Dialog System Technology Challenge (DSTC11)}},
  author    = {Hagen Soltau and others},
  year      = {2023},
  booktitle = {Interspeech 2023},
  pages     = {4668--4672},
  doi       = {10.21437/Interspeech.2023-1037},
  issn      = {2958-1796},
}

@inproceedings{radford2023robust,
  title={Robust speech recognition via large-scale weak supervision},
  author={Radford, Alec and Kim, Jong Wook and Xu, Tao and Brockman, Greg and McLeavey, Christine and Sutskever, Ilya},
  booktitle={International conference on machine learning},
  pages={28492--28518},
  year={2023},
  organization={PMLR}
}

@inproceedings{wang2021voxpopuli,
  title={VoxPopuli: A Large-Scale Multilingual Speech Corpus for Representation Learning, Semi-Supervised Learning and Interpretation},
  author={Wang, Changhan and others},
  booktitle={ACL 2021-59th Annual Meeting of the Association for Computational Linguistics},
  year={2021}
}

@inproceedings{ardila-etal-2020-common,
    title = "Common Voice: A Massively-Multilingual Speech Corpus",
    author = "Ardila, Rosana  and others",
    booktitle = "Proceedings of the Twelfth Language Resources and Evaluation Conference",
    month = may,
    year = "2020",
    address = "Marseille, France",
    publisher = "European Language Resources Association",
    url = "https://aclanthology.org/2020.lrec-1.520/",
    pages = "4218--4222",
    language = "eng",
    ISBN = "979-10-95546-34-4"
}

@INPROCEEDINGS{panayatov2015librispeech,
  author={Panayotov, Vassil and Chen, Guoguo and Povey, Daniel and Khudanpur, Sanjeev},
  booktitle={ICASSP}, 
  title={{Librispeech: An ASR corpus based on public domain audio books}}, 
  year={2015},
  volume={},
  number={},
  pages={5206-5210},
  doi={10.1109/ICASSP.2015.7178964}}

@inproceedings{cieri2004fisher,
  title={{The Fisher corpus: A resource for the next generations of speech-to-text.}},
  author={Cieri, Christopher and Miller, David and Walker, Kevin},
  booktitle={LREC},
  volume={4},
  pages={69--71},
  year={2004}
}

@inproceedings{
loshchilov2018decoupled,
title={Decoupled Weight Decay Regularization},
author={Ilya Loshchilov and Frank Hutter},
booktitle={International Conference on Learning Representations},
year={2019},
url={https://openreview.net/forum?id=Bkg6RiCqY7},
}

@inproceedings{
hu2022lora,
title={Lo{RA}: Low-Rank Adaptation of Large Language Models},
author={Edward J Hu and others},
booktitle={International Conference on Learning Representations},
year={2022},
url={https://openreview.net/forum?id=nZeVKeeFYf9}
}

@inproceedings{nekvinda-dusek-2021-shades,
    title = "Shades of {BLEU}, Flavours of Success: The Case of {M}ulti{WOZ}",
    author = "Nekvinda, Tom{\'a}{\v{s}}  and
      Du{\v{s}}ek, Ond{\v{r}}ej",
    booktitle = "Proceedings of the First Workshop on Natural Language Generation, Evaluation, and Metrics (GEM)",
    month = aug,
    year = "2021",
    address = "Online",
    publisher = "Association for Computational Linguistics",
    url = "https://aclanthology.org/2021.gem-1.4/",
    doi = "10.18653/v1/2021.gem-1.4",
    pages = "34--46"
}

@inproceedings{zhang-etal-2024-dialogstudio,
    title = "{D}ialog{S}tudio: Towards Richest and Most Diverse Unified Dataset Collection for Conversational {AI}",
    author = "Zhang, Jianguo  and
      others",
    booktitle = "Findings of the Association for Computational Linguistics: EACL 2024",
    month = mar,
    year = "2024",
    address = "St. Julian{'}s, Malta",
    publisher = "Association for Computational Linguistics",
    url = "https://aclanthology.org/2024.findings-eacl.152/",
    pages = "2299--2315"
}

@inproceedings{yoon-etal-2023-adapting,
    title = "Adapting Text-based Dialogue State Tracker for Spoken Dialogues",
    author = "Yoon, Jaeseok  and
      Hwang, Seunghyun  and
      Ran, Han  and
      Bang, Jeong-Uk  and
      Kim, Kee-Eung",
    booktitle = "Proceedings of the Eleventh Dialog System Technology Challenge",
    month = sep,
    year = "2023",
    address = "Prague, Czech Republic",
    publisher = "Association for Computational Linguistics",
    url = "https://aclanthology.org/2023.dstc-1.10/",
    pages = "81--88"
}

@inproceedings{jiang-etal-2023-speech,
    title = "Speech-Aware Multi-Domain Dialogue State Generation with {ASR} Error Correction Modules",
    author = "Jiang, Ridong  and
      others",
    booktitle = "Proceedings of the Eleventh Dialog System Technology Challenge",
    month = sep,
    year = "2023",
    address = "Prague, Czech Republic",
    publisher = "Association for Computational Linguistics",
    url = "https://aclanthology.org/2023.dstc-1.13/",
    pages = "105--112"
}

@inproceedings{sunder23_interspeech,
  title     = {{ConvKT: Conversation-Level Knowledge Transfer for Context Aware End-to-End Spoken Language Understanding}},
  author    = {Vishal Sunder and Eric Fosler-Lussier and Samuel Thomas and Hong-Kwang J Kuo and Brian Kingsbury},
  year      = {2023},
  booktitle = {Interspeech 2023},
  pages     = {1129--1133},
  doi       = {10.21437/Interspeech.2023-2018},
  issn      = {2958-1796},
}

@inproceedings{ganhotra21_interspeech,
  title     = {Integrating Dialog History into End-to-End Spoken Language Understanding Systems},
  author    = {Jatin Ganhotra and others},
  year      = {2021},
  booktitle = {Interspeech 2021},
  pages     = {1254--1258},
  doi       = {10.21437/Interspeech.2021-1460},
  issn      = {2958-1796},
}

@inproceedings{wang2024retrieval,
  title={Retrieval augmented end-to-end spoken dialog models},
  author={Wang, Mingqiu and others},
  booktitle={ICASSP},
  pages={12056--12060},
  year={2024},
  organization={IEEE}
}

@article{thulke2021adapting,
  title={Adapting document-grounded dialog systems to spoken conversations using data augmentation and a noisy channel model},
  author={Thulke, David and Daheim, Nico and Dugast, Christian and Ney, Hermann},
  journal={arXiv preprint arXiv:2112.08844},
  year={2021}
}

@article{tian2021tod,
  title={{TOD-DA: Towards Boosting the Robustness of Task-oriented Dialogue Modeling on Spoken Conversations}},
  author={Tian, Xin and others},
  journal={arXiv preprint arXiv:2112.12441},
  year={2021}
}

@inproceedings{chen22r_interspeech,
  title     = {{MAESTRO: Matched Speech Text Representations through Modality Matching}},
  author    = {{Zhehuai Chen and others}},
  year      = {{2022}},
  booktitle = {{Interspeech 2022}},
  pages     = {{4093--4097}},
  doi       = {{10.21437/Interspeech.2022-10937}},
  issn      = {{2958-1796}},
}

@inproceedings{yusuf2022usted,
  title={{USTED: Improving ASR with a unified speech and text encoder-decoder}},
  author={Yusuf, Bolaji and Gandhe, Ankur and Sokolov, Alex},
  booktitle={ICASSP},
  pages={8297--8301},
  year={2022},
  organization={IEEE}
}

@article{yusuf2024written,
  title={{Written Term Detection Improves Spoken Term Detection}},
  author={Yusuf, Bolaji and Sara{\c{c}}lar, Murat},
  journal={IEEE/ACM Transactions on Audio, Speech, and Language Processing},
  volume={32},
  pages={3213--3223},
  year={2024},
  publisher={IEEE}
}

@inproceedings{dao25_interspeech,
  title     = {{Speechless: Speech Instruction Training Without Speech for Low Resource Languages}},
  year      = {2025},
  booktitle = {{Interspeech 2025}},
  pages     = {3239--3243},
  doi       = {10.21437/Interspeech.2025-1292},
  issn      = {2958-1796},
}

@inproceedings{thomas2022towards,
  title={Towards reducing the need for speech training data to build spoken language understanding systems},
  author={Thomas, Samuel and Kuo, Hong-Kwang J and Kingsbury, Brian and Saon, George},
  booktitle={ICASSP 2022-2022 IEEE International Conference on Acoustics, Speech and Signal Processing (ICASSP)},
  pages={7932--7936},
  year={2022},
  organization={IEEE}
}

@inproceedings{tang-etal-2021-improving,
    title = "Improving Speech Translation by Understanding and Learning from the Auxiliary Text Translation Task",
    author = "Tang, Yun  and
      Pino, Juan  and
      Li, Xian  and
      Wang, Changhan  and
      Genzel, Dmitriy",
    booktitle={59th Annual Meeting of the Association for Computational Linguistics},
    month = aug,
    year = "2021",
    address = "Online",
    publisher = "Association for Computational Linguistics",
    url = "https://aclanthology.org/2021.acl-long.328/",
    doi = "10.18653/v1/2021.acl-long.328",
    pages = "4252--4261"
}

\end{document}